\documentclass[runningheads]{llncs}
\usepackage[T1]{fontenc}
\usepackage{graphicx}

\usepackage{epsfig}
\usepackage{amsmath}
\usepackage{amssymb}
\usepackage{subcaption}
\usepackage{booktabs}
\usepackage{braket}
\usepackage{xcolor}
\usepackage{float}
\usepackage{wrapfig}

\usepackage{algorithm}
\usepackage{algorithmic}

\newtheorem{assumption}{Assumption}

\begin{document}
%

\title{Causality-Aware Transformer Networks for Robotic Navigation}

\titlerunning{Causality-Aware Transformer Networks for Robotic Navigation}
%
\author{Ruoyu Wang \inst{1} \and
Yao Liu \inst{2} \and 
Yuanjiang Cao \inst{2} \and
Lina Yao \inst{1,3}}
%
\authorrunning{Wang et al.}
%
\institute{University of New South Wales \and 
Macquarie University \and
Commonwealth Scientific and Industrial Research Organisation, Australia \email{ruoyu.wang5@unsw.edu.au}, \email{y.liu@mq.edu.au}, \email{yuanjiang.cao@mq.edu.au}, \email{lina.yao@unsw.edu.au}}

\maketitle              
\begin{abstract}
Current research in Visual Navigation reveals opportunities for improvement. First, the direct adoption of RNNs and Transformers often overlooks the specific differences between Embodied AI and traditional sequential data modelling, potentially limiting its performance in Embodied AI tasks. Second, the reliance on task-specific configurations, such as pre-trained modules and dataset-specific logic, compromises the generalizability of these methods. We address these constraints by initially exploring the unique differences between Navigation tasks and other sequential data tasks through the lens of Causality, presenting a causal framework to elucidate the inadequacies of conventional sequential methods for Navigation. By leveraging this causal perspective, we propose Causality-Aware Transformer (CAT) Networks for Navigation, featuring a Causal Understanding Module to enhance the models's Environmental Understanding capability. Meanwhile, our method is devoid of task-specific inductive biases and can be trained in an End-to-End manner, which enhances the method's generalizability across various contexts. Empirical evaluations demonstrate that our methodology consistently surpasses benchmark performances across a spectrum of settings, tasks and simulation environments. Extensive ablation studies reveal that the performance gains can be attributed to the Causal Understanding Module, which demonstrates effectiveness and efficiency in both Reinforcement Learning and Supervised Learning settings.

\end{abstract}
\section{Introduction}
Navigation is a fundamental task in the research of Embodied AI, and the methods for Navigation can be broadly grouped into {\it Supervised Learning methods} and {\it Reinforcement Learning methods}, distinguished by the nature of their training processes \cite{francis2022core,smith2005development,duan2022survey}. Supervision in Navigation generally refers to a demonstration of a possible solution to a problem, and the methods are typically focused on matching the behaviour of their demonstrator. In Reinforcement Learning (RL) settings, an agent learns the policy by interactions with an environment. In both settings, while some methods have been proposed in recent years, many of them share similar limitations.

First, many methods involve the incorporation of task-specific inductive bias thus lack of generalizability \cite{duan2022survey}. For example, they usually require hand-crafted logics \cite{zheng2022jarvis,min2021film,inoue2022prompter}, pre-trained modules to construct semantic maps \cite{luo2022stubborn,min2021film,zhang2021hierarchical}, additional large dataset \cite{suglia2021embodied,deitke2022️}, or multi-stage training \cite{blukis2022persistent,pashevich2021episodic,inoue2022prompter}. These limitations render them deficient in terms of generalizability and reproducibility, precluding their extension to alternative datasets or tasks.

Second, most existing methods directly adopt the traditional sequential data modeling methods such as RNN or Transformer as a core module. While employing these strategies from analogous tasks involving sequential data might seem expedient, it is crucial to acknowledge that Navigation tasks exhibit distinct characteristics from a causal perspective (Section~\ref{method_motivation}). However, existing methods haven't adequately engaged with these distinctions thus limiting their effectiveness on these tasks.


While a few existing works \cite{khandelwal2022simple} tackled the generalizability problem by proposing an End-to-End method without task-specific architectures and inductive biases, the second weakness introduced above remains unaddressed. Therefore, we introduce a novel method to address these identified shortcomings in this paper.

Firstly, to enhance the generalizability and reproducibility of the approaches, we present an end-to-end method for visual navigation. This framework eschews any task-specific architecture and inductive biases, aligning with the principles demonstrated in EmbCLIP \cite{khandelwal2022simple}. Secondly, we conduct a thorough analysis of the disparities between navigation tasks and other types of sequential tasks through the lens of causality. By encapsulating these distinctions within a causal framework, we elucidate the inadequacies of conventional sequential methods, such as RNNs and Transformers, in addressing these tasks. Consequently, we propose an innovative solution incorporating a Causal Understanding Module, designed to significantly enhance model performance in these contexts. In a nutshell, our contributions are three-folded:

\begin{itemize}
    \item We propose Causality-Aware Transformer (CAT), a novel End-to-End framework for Navigation tasks without task-specific inductive bias, and experimentally show that it outperforms the baseline methods by a significant margin across various tasks and simulators.
    
    \item We introduce a causal framework for Navigation tasks, 
    offering a comprehensive rationale for the limitations in existing methods that potentially hinder agent performance. Without incurring additional computational cost, we design a Causal Understanding Module to substantially enhance the methods' effectiveness and efficiency, demonstrating the framework's ability to optimize performance through a more nuanced understanding of causality.
    
    \item We conduct comprehensive ablation studies and demonstrate the efficiency, effectiveness and necessity of the Causal Understanding Module. Besides, while we primarily focus on the tasks in the Reinforcement Learning setting, experiments show that the proposed Causal Understanding Module is also generally beneficial in the Supervised Learning setting.
\end{itemize}

\begin{wrapfigure}{R}{0.6\textwidth}
    \vspace{-25pt}
    \centering
    \includegraphics[width=0.6\textwidth]{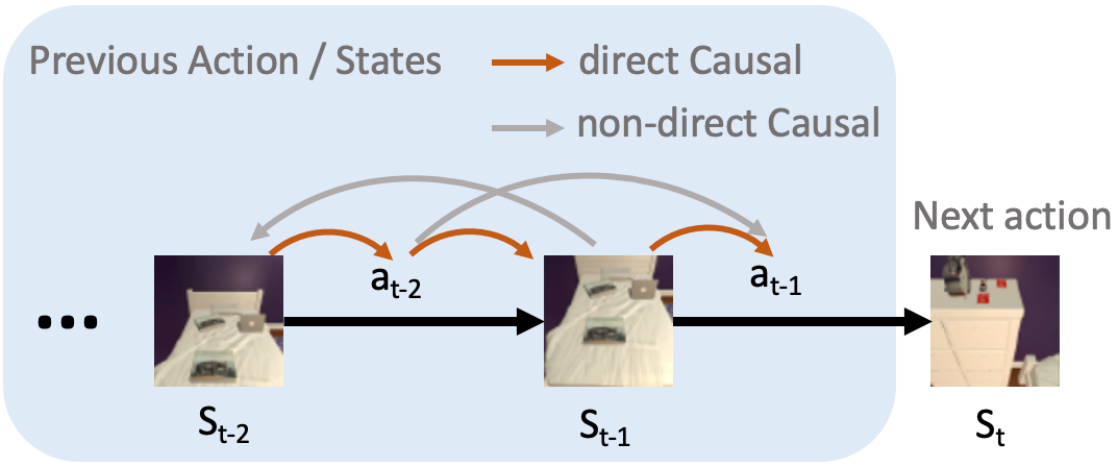}
    \caption{Our method encourages the model to understand the environment by highlighting the direct causal relationships and diminishing the non-direct causal associations.}
    \label{fig:direct_causal}
    \vspace{-20pt}
\end{wrapfigure}

\section{Background}
\label{method}

\subsection{Motivation}
\label{method_motivation}
Across diverse settings and paradigms within Embodied AI tasks, many necessitate the modelling of sequential data accumulated from preceding time steps. Predominantly, research endeavours utilize Recurrent Neural Networks (RNNs) or Transformers for this purpose, given their established efficacy in handling sequential data. Nevertheless, these methods were primarily proposed for other scenarios such as NLP, our question is, are these methods well-suited for the Navigation tasks?

We argue that directly adopting methods such as RNNs or Transformers to navigation tasks may result in specific limitations. This is due to the intrinsic differences between navigation tasks and conventional sequential data modelling tasks such as NLP, particularly when viewed through the framework of Causality. For instance, if we consider the following sentence in NLP:

\begin{center}
    \textit{The doctor asked the nurse a question, ``...?'',  She said, .... }
\end{center}
It is obvious that the token \textit{She} refers to \textit{the nurse}, which was mentioned several time steps earlier. This indicates that direct causal relationships in such scenarios \textbf{can be long-term or short-term}, and do not adhere to a universal pattern.

In Navigation, however, based on our empirical understanding of the physical world, the direct causal relationship \textbf{can only be one-step} (Figure~\ref{fig:direct_causal}), because it is generally believed that the transition from one state to another is attributed to the actions undertaken in the interim. For example, the current view of the agent and the action performed in that state serve as the direct determinants of its subsequent view. While previous views also affect future states, these relationships are \textbf{indirect} and \textbf{must be mediated by the current state}. 

This key distinction underscores why RNNs or Transformers may not be ideally suited for Navigation tasks, as they are inherently designed to capture long-term associations across time steps. While this feature is advantageous for tasks that exhibit long-term causal connections, it proves counterproductive for navigation, where such connections are weak. Thus, architectural modifications are required to address these limitations.

\subsection{Motivation Formulation from a Causal Perspective}
\label{method_causal_framework}

We formalize the motivation introduced above into a causal framework. First, based on the nature of the Navigation introduced above, we make the following assumptions, which closely align with our empirical observations in the real world, as elaborated in Section~\ref{method_motivation}.

\begin{assumption}
    At any given time step $t$, the state $S_{t}$ and the action $a_{t}$ are the only direct causal parent influencing the subsequent state at time step $t+1$, denoted as $S_{t+1}$.
    \label{assumption:state_action}
\end{assumption}

\begin{assumption}
    At any given time step $t$, the state $S_{t}$ and the Objective are the only causal parents of the action $a_{t}$.
    \label{assumption:state_goal_action}
\end{assumption}

\textbf{Differences with Markov Property} While the intuition of these assumptions may appear similar to the Markov Property \cite{puterman1990markov}, they are distinct concepts from three perspectives: 1) Our assumptions pertain to causality, whereas the Markov property concerns transitional probability. 2) The Markov property underpins problem formulations such as Markov Decision Processes (MDPs), rather than serving as a specific modeling technique. However, our focus is on the practical aspects, and we discuss the appropriate architecture for modelling the scenario. 3) Our method is broadly applicable in both Reinforcement Learning and Supervised Learning contexts, thus not restricted to MDPs. 

\begin{figure*}[t]
  \centering
  \includegraphics[width=0.9\linewidth]{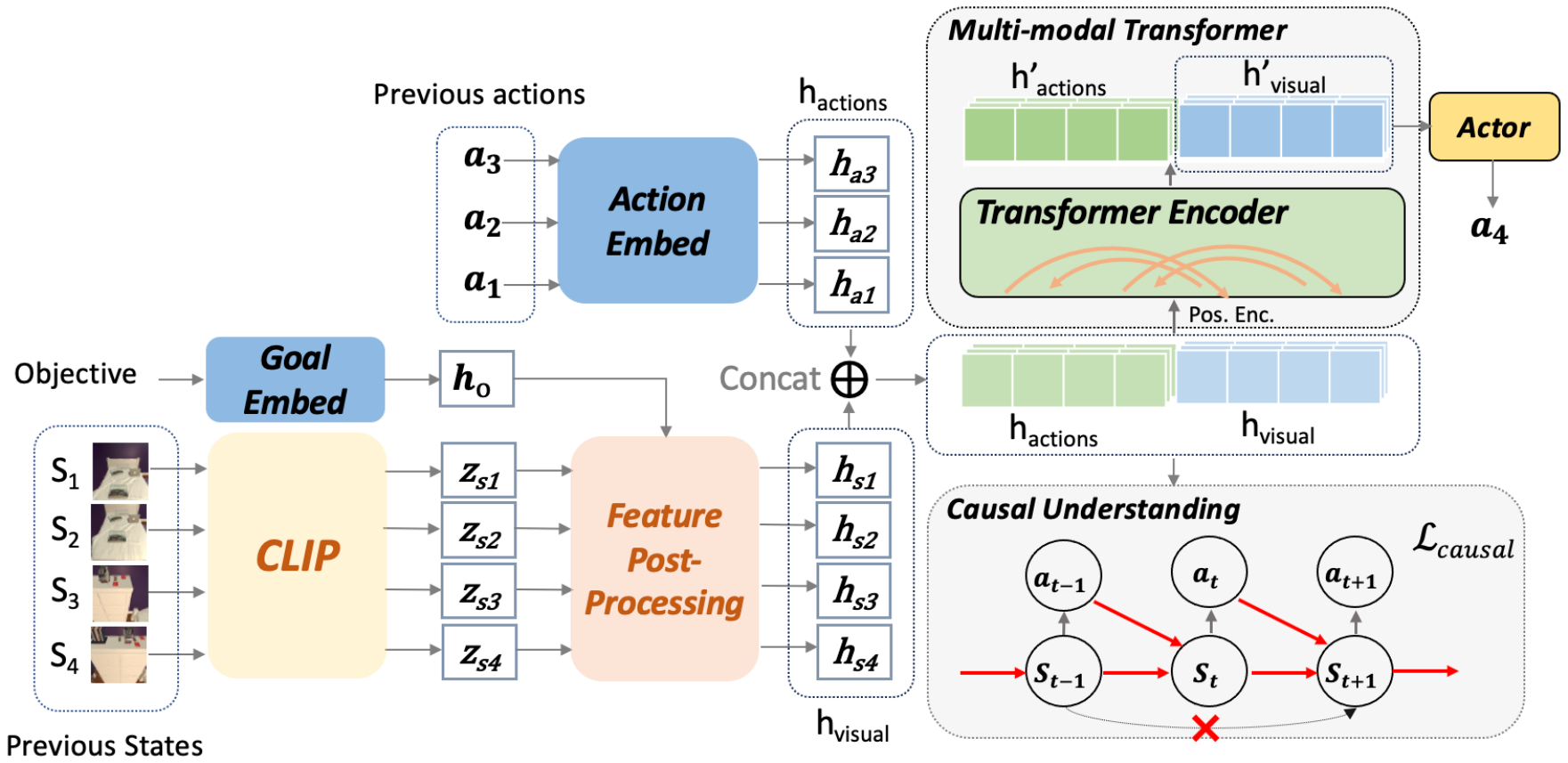}
  \caption{The framework of our proposed method. First, we process the visual states by the CLIP vision model and process the objective and previous actions by simple Embedding modules. After a tunable Feature Post-Processing layer, we concatenate the features of the states and the actions and process the features with a Transformer Encoder. Finally, the Actor layer takes the post-processed visual features as input to predict the action at the current time step.}
  \label{fig:framework}
  \vspace{-15pt}
\end{figure*}

We depict these causal assumptions in a causal graph, as illustrated in the enlarged Causal Understanding Module in Figure~\ref{fig:framework}, where the edge $S_{t-1} \rightarrow S_{t} \leftarrow a_{t-1}$ corresponds to the Assumption~\ref{assumption:state_action}, and the edge $S_{t} \rightarrow a_{t}$ corresponds to the Assumption~\ref{assumption:state_goal_action}. Further, we derive Proposition~\ref{proposition:no_long_term} based on these assumptions, which suggests that there exist NO direct causal relationships between $S_{t-1}$ and $S_{t+1}$, so $S_{t-1} \rightarrow S_{t+1}$ is marked by a light dotted line, indicating the causation between them is indirect and weak.
\begin{proposition}
    At any given time step $t$, and for any integer $\delta \geq 2$, there exist no direct causal relationships between $S_{t}$ and $S_{t-\delta}$, the causal relationships between states $S_{t}$ and $S_{t-\delta}$ are indirect and must be mediated by states $S_{t'}$ for all $t'$ where $t-\delta \leq t' \leq t$.
    \label{proposition:no_long_term}
\end{proposition}
As discussed in Section~\ref{method_motivation}, we identify Proposition~\ref{proposition:no_long_term} as a crucial distinction between Navigation tasks and other sequential data tasks such as NLP. Note that we are not denying the existence of long-term associations between $S_{t}$ and $S_{t-\delta}$. Instead, we argue that these associations should be mediated by intermediate states, making them weaker than short-term associations driven by direct causation. Therefore, to encourage the direct causal relationship $S_{t-1} \rightarrow S_{t}$ and $a_{t-1} \rightarrow S_{t}$ \textbf{stand out} from the associations in other forms, we introduce our method in Section~\ref{method_transformer}.

\section{Causality-Aware Transformer Networks}
\label{method_transformer}

The architecture of our proposed methodology is depicted in Figure~\ref{fig:framework} and encompasses multiple components: {\bf Visual Encoder}, {\bf Goal/Action Embedding}, {\bf Feature Post-Processing}, {\bf Multi-modal Transformer}, and the {\bf Causal Understanding Module}. The functions and operations of these individual modules are delineated in the subsequent sections.

\textbf{Visual Encoder} We fine-tune the pre-trained CLIP ResNet-50 model \cite{radford2021learning} as the Visual Encoder to convert the visual signal $S_{i}$ into features $z_{s_{i}}$ because CLIP is proven to be effective for the Embodied AI tasks \cite{khandelwal2022simple}.

\textbf{Goal \& Action Embedding} Since the number of possible targets and actions in each task is finite, we utilize a simple Embedding layer, which functions as a lookup table that stores embeddings of a fixed dictionary, to transform the Objective and Actions $a_{i}$ such as \textit{Find the laptop}, \textit{Move Forward}, and \textit{Turn Left} into features $h_{o}$ and $h_{a_{i}}$. 

\textbf{Feature Post-Processing} After obtaining the representations of the visual signal of each state $z_{s_{i}}$ and the representation of the goal $h_{o}$, we pass these features to a Feature Post-Processing Module, which is designed to integrate objective-related information into the visual representation of each state serves to reinforce the causal relationship between the objective and the subsequent action (Assumption~\ref{assumption:state_goal_action}). Consequently, the Feature Post-Processing Module generates a distinct set of features $h_{s_{i}}$ for each state.

\textbf{Multi-modal Transformer} All the features obtained in the earlier steps are passed into a Multi-modal Transformer Module, as illustrated in Figure~\ref{fig:framework}. We use $h_{visual}$ and $h_{action}$ to denote the sequence of ${h_{v_{i}}}$ and ${h_{a_{i}}}$ for simplicity.

First, we concatenate $h_{visual}$ and $h_{action}$ along the direction of the time step, as demonstrated by Equation~\ref{eq:concat}, and apply the Positional Encoding on the concatenated features $h_{concat}$. Then, the features will be passed to the transformer encoder \cite{vaswani2017attention}. In particular, we use a causal attention mechanism that prevents visual and action embeddings from attending to subsequent time steps. Consequently, the transformer produces a set of updated features $h'_{concat}$.
\begin{equation}
    h_{concat} = \left[ h_{visual}; h_{actions} \right]
    \label{eq:concat}
\end{equation}
Similarly, the output of the transformer contains the updated features of visual states and previous actions, as illustrated by Equation~\ref{eq:output_concat}. Then, we take the updated visual features $h'_{visual}$ and feed them to an Actor layer, which predicts the proper action to be taken at the given state, as illustrated by Equation~\ref{eq:actor}. Finally, the agent will interact with the environment through $a_{t}$ to get the visual signal of the next state, then the framework proceed to the next time step.

\begin{equation}
    h'_{concat} = \left[ h'_{visual}; h'_{actions} \right]
    \label{eq:output_concat}
\end{equation}
\begin{equation}
    a_{t} = Actor(h'_{visual})
    \label{eq:actor}
\end{equation}

\textbf{Causal Understanding Module} Following the idea in Assumption~\ref{assumption:state_action}, Assumption~\ref{assumption:state_goal_action} and Proposition~\ref{proposition:no_long_term}, we use a Causal Understanding Module to encourage the one-step direct causation to stand out from the associations in other forms. In particular, we claim that for any time step $t$, the state $S_{t}$ and the corresponding action $a_{t}$ should be the \textit{only direct causes} of the next state $S_{t+1}$. And any states in earlier time steps only hold indirect causation with $S_{t+1}$, thus the association between them should be weaker than the direct causation between adjacent states. To align the model architecture with these assumptions, we aim to make the direct causations stand out from other forms of associations. Therefore, we introduce a Causal Understanding Module, which predicts the representation of the next state $S_{t+1}$ by taking the representation of $S_{t}$ and $a_{t}$ as input. Ideally, a model demonstrating accurate predictions on this auxiliary task signifies a thorough understanding and effective encoding of the underlying environment. Mathematically, the Causal Understanding Module is trained by minimizing the Causal Loss as defined in Equation~\ref{eq:loss_causal}.
\begin{equation}
    \mathcal{L}_{causal}(\theta) = \mathbb{E}_{t} \left[ (Causal(h_{v_{t}}, h_{a_{t}}) - h_{v_{t+1}} )^2 \right]    
    \label{eq:loss_causal}
\end{equation}

\textbf{Training Process} For experiments in the Reinforcement Learning setting, we train our model with Proximal Policy Optimization (PPO) \cite{schulman2017proximal}. So the overall objective of our framework becomes Equation~\ref{eq:loss_overall}, where $\mathcal{L}_{PPO}$ denotes the original objective of PPO \cite{schulman2017proximal} which is to be maximized, $\mathcal{L}_{causal}$ is our proposed Causal Loss which is to be minimized, so we subtract the causal loss from the PPO objective.
\begin{equation}
    \mathcal{L}_{total}(\theta) = \mathcal{L}_{PPO} - \alpha \mathcal{L}_{causal}
    \label{eq:loss_overall}
\end{equation}
We also conduct experiments in the Supervised Learning setting, where the training process differs from that in the Reinforcement Learning setting. This will be elaborated in Section~\ref{exp_other_setting}.


\section{Experiments}
\label{exp}
In Section~\ref{exp_setting}-\ref{exp_rst}, we evaluate the performance of our method in the Reinforcement Learning setting. In Section~\ref{exp_ablation}, we conduct ablation studies and found the Causal Understanding Module contributes most of the performance gain. In Section~\ref{exp_other_setting}, we demonstrate that the Causal Understanding Module is also generally applicable and effective in the Supervised Learning setting.

\subsection{Experiment Setting}
\label{exp_setting}
\subsubsection {Task Descriptions}
We evaluate our method over three tasks in the Reinforcement Learning setting:
(1) \textbf{Object Navigation in RoboTHOR} \cite{deitke2020robothor}, which requires an agent to navigate through its environment and find an object of a given category. For example, \textit{Find an apple}. The task consists of 12 possible goal object categories, and the agent is allowed to \textit{MoveAhead}, \textit{RotateRight}, \textit{RotateLeft}, \textit{LookUp}, and \textit{LookDown}. The agent is considered to have completed the task if it takes a special \textit{Stop} action and one goal object category is visible within 1 meter of the agent. 
(2) \textbf{Object Navigation in Habitat}, which is defined similarly to the task in RoboTHOR, but Habitat has 21 objects and does not require the agent to be looking at a target object to succeed. Habitat uses scenes from the MatterPort3D \cite{chang2017matterport3d} dataset of real-world indoor spaces. 
(3) \textbf{Point Navigation in Habitat}, which requires an agent to navigate from a random initial position to polar goal coordinates. For example, \textit{Navigate to (X, Y)}. The agent is allowed to do three actions, which include \textit{MoveAhead}, \textit{RotateRight}, and \textit{RotateLeft}. The agent should perform a special \textit{Done} action when it reaches its goal coordinates. We train within the Gibson Database \cite{xia2018gibson}.

\subsubsection{Evaluation Metrics}
Various metrics are employed to assess an agent's performance across different tasks. Following the previous works, we evaluate the performance of an agent on Object Navigation in RoboTHOR by Success Rate (SR) and Success weighted by Path Length (SPL).
Success Rate measures the frequency with which an agent successfully completes a task, while Success weighted by Path Length (SPL) takes into account the length of the path traversed by the agent to accomplish the task. On the other hand, for the two tasks in Habitat, apart from SR and SPL, we also evaluate the agent's performance on Goal Distance (GD), which quantifies the agent's proximity to the goal upon task completion. In general, an agent is considered to be better if it achieves a higher SR and SPL, or lower GD.

\subsubsection{Baselines}

(1) We compare our method with EmbCLIP, because both methods do not contain any task-specific designs and can be trained in an End-to-End manner, thus lying in the same category and can be compared directly. For both methods, we train the model for 20M steps. And during the training process, we select the checkpoint with the highest SR for evaluation and comparison.

(2) We also compare our method with baseline methods specifically designed for each task. However, as discussed earlier, it is worth noting that these methods are not directly comparable with us because they are not in the same setting as our method. For example, these methods employ either task-specific hand-crafted logic, undergo training across multiple stages, or leverage extensive offline datasets to facilitate the training process. But for a comprehensive understanding of the performance of our method, we report the validation results of these methods as a reference.
Specifically, we compare our method with Action Boost, RGB+D, ICT-ISIA, and ProcTHOR \cite{deitke2022️} on RoboTHOR ObjNav; and compare with Stubborn \cite{luo2022stubborn}, TreasureHunt \cite{maksymets2021thda}, Habitat on Web (IL-HD) \cite{ramrakhya2022habitat}, Red Rabbit \cite{ye2021auxiliary}, PIRLNav\cite{ramrakhya2023pirlnav} and RIM on Habitat ObjNav; and compare with DD-PPO \cite{wijmans2019dd}, Monocular Predicted Depth, Arnold and SRK AI on Habitat PointNav.

\begin{figure}[t]

  \centering
  \includegraphics[width=1.0\textwidth]{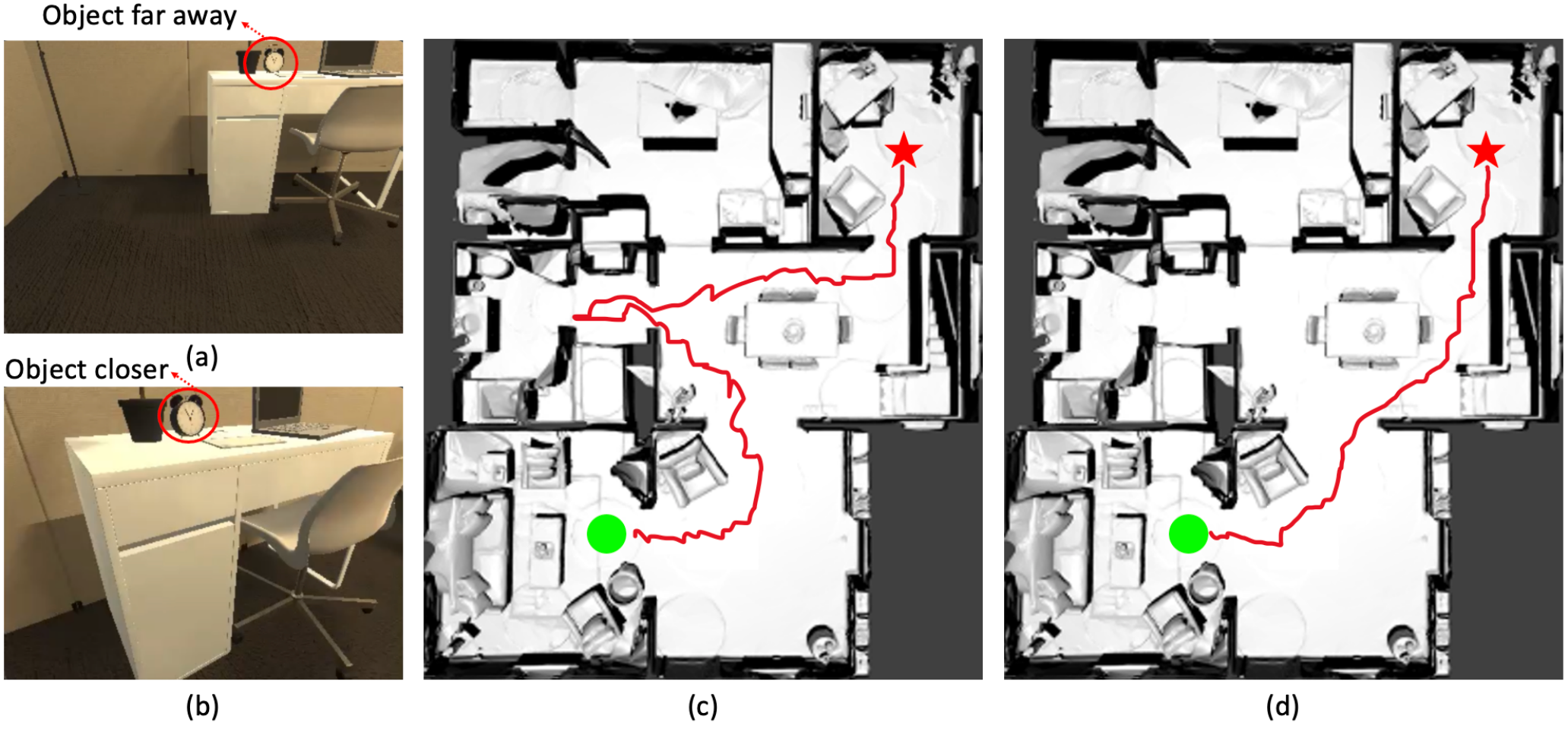}
  \caption{Effect of our method. (a)-(b) Comparison on RoboTHOR ObjNav \textit{Find an AlarmClock} task. Our method encourages the agent to stop at a spot closer to the goal object, thus benefiting the performance (c)-(d) Comparison on Habitat PointNav task. Our method allows the agent to directly navigate to the target point by choosing an optimal route.}
  \label{fig:case_eg}
  \vspace{-10pt}
\end{figure}

\subsubsection{Environment and Parameters} 
For all the experiments, we trained our method for 20M steps. and select the checkpoint with the highest Success Rate (SR) for evaluation on other metrics. Regarding the model architecture, we use an Embedding layer for Action and Objective Embedding and a linear layer for the Feature Post-Processing, the Casual Understanding and the Actor Module. The Multi-modal Transformer Encoder is a single-layer transformer encoder with 4 self-attention heads and the dimension for each self-attention head is 392 thus the total dimension is 1568. The episode length is 128. The weight $\alpha$ for Causal Loss is 1. We train our framework using Adam \cite{kingma2014adam} with a learning rate of 1e-4, and schedule the learning rate to linearly Decay to 0 when the training is finished. Our code is based on AllenAct \cite{weihs2020allenact}, a learning framework designed for Embodied-AI research. Any settings not specified here remain the same as \cite{khandelwal2022simple}. 

\subsection{Results}
\label{exp_rst}

    

\begin{table}
\caption{Result on RoboTHOR ObjNav (Left) and Habitat PointNav (Right). Our method outperforms the baseline significantly and also performs better than the methods designed with stronger assumptions. SR and GD evaluations for some previous works are not provided as they are not publicly available.}
\small
\begin{minipage}{.45\linewidth}
    \centering
    \label{tab:rst_objnav_robothor}
    \setlength{\tabcolsep}{6pt}
    \medskip
    \begin{tabular}{l | c c c c}
    \toprule
    & SPL $\uparrow$ & SR $\uparrow$ \\
    \midrule
    CAT (Ours) & \textbf{0.31} & \textbf{0.69} \\
    Emb-CLIP & 0.14 & 0.31 \\
    
    \midrule
    ProcTHOR  & 0.27 & 0.66 \\
    Action Boost & 0.17 & 0.37 \\
    RGB+D ResNet18 & 0.17 & 0.35 \\
    ICT-ISIA & 0.18 & 0.38 \\
    \bottomrule
  \end{tabular}
\end{minipage}\hfill
\begin{minipage}{.52\linewidth}
    \centering
    \setlength{\tabcolsep}{6pt}
    \label{tab:rst_pointnav_habitat}
    \medskip
    \begin{tabular}{l | c c c}
    \toprule
    & SPL $\uparrow$ & SR $\uparrow$ & GD $\downarrow$ \\
    \midrule
    CAT (Ours) & \textbf{0.93} & \textbf{0.98} & \textbf{0.34} \\
    Emb-CLIP & 0.80 & 0.92 & 0.48 \\
    \midrule
    DD-PPO & 0.89 & - & - \\
    MPD & 0.85 & - & - \\
    Arnold & 0.70 & - & - \\
    SRK AI Lab & 0.66 & - & - \\
    \bottomrule
  \end{tabular}
\end{minipage}
\vspace{-10pt}

\end{table}

We provide the results of our experiments for each task in Table~\ref{tab:rst_objnav_robothor} and Table~\ref{tab:rst_objnav_habitat}. As introduced in Section~\ref{exp_setting}, we compare our method with two types of baselines, thus our findings similarly fall into two aspects:

(1) Compared with EmbCLIP, our method outperforms the baseline \textbf{significantly} on all the tasks across all evaluation metrics, as illustrated in Table~\ref{tab:rst_objnav_robothor}-\ref{tab:rst_objnav_habitat}. This shows the effectiveness of our method. In particular, our method achieves more than a doubling of EmbCLIP on SPL and SR, in RoboTHOR Object Navigation and Habitat Object Navigation tasks.

(2) Compared with other methods with stronger assumptions and inductive bias, we observe that our method also achieves \textbf{better} results than these methods. While a few of these methods achieve similar performance to our method on some evaluation metrics, these method lacks generalizability and reproducibility due to various reasons, as elaborated earlier. For example, in Table~\ref{tab:rst_objnav_robothor}, while ProcTHOR\cite{deitke2022️} also performs well, it requires extremely large offline data to train a large model first, and then fine-tune the pre-trained model as a second stage. Similarly, in Table~\ref{tab:rst_objnav_habitat}, while RIM and PIRLNav \cite{ramrakhya2023pirlnav} achieve a similar result to us, they also require multi-stage training and task-specific designs.

\begin{wraptable}{R}{6cm}
  \setlength{\tabcolsep}{5pt}
  \centering
  \normalsize
  \vspace{-15pt}
  \caption{Results on Habitat ObjNav.}
  \begin{tabular}{l | c c c}
    \toprule
    & SPL & SR & GD \\
    \midrule
    CAT (Ours) & \textbf{0.16} & \textbf{0.41} & \textbf{6.76} \\
    Emb-CLIP & 0.07 & 0.15 & 7.13 \\
    
    \midrule
    RIM & 0.15 & 0.37 & 6.80 \\
    PIRLNav & 0.14 & 0.35 & 6.95 \\
    Stubborn & 0.10 & 0.22 & 9.17 \\
    TreasureHunt & 0.09 & 0.21 & 9.20 \\
    Habitat on Web & 0.08 & 0.24 & 7.88 \\
    Red Rabbit & 0.06 & 0.24 & 9.15 \\
    \bottomrule
  \end{tabular}
  \vspace{-15pt}
  \label{tab:rst_objnav_habitat}
\end{wraptable}

Besides, we conducted some \textbf{case studies} to investigate the benefits of implementing our method. In Figure~\ref{fig:case_eg}a-b, we tested our method and EmbCLIP with the task ``Find an AlarmClock'' in RoboTHOR ObjNav, and found that our method encouraged the agent to stop at a spot closer to the objective, thus benefiting the performance. In Figure~\ref{fig:case_eg}c-d, we test the methods in Habitat PointNav with the \textit{Nuevo} scenario, where the green dot denotes the starting point and the red star denotes the target point. We found that our agent can directly navigate to the target point with an optimal route, while EmbCLIP failed to find the shortest path. We speculate that this is because our method enables the agent to understand the environment more efficiently, so the agent can navigate to a place that was less explored before.

In summary, both quantitative results and the qualitative sample check demonstrated the effectiveness and generalizability of our proposed method.

\subsection{Ablation Studies}
\label{exp_ablation}

Our proposed method differs from EmbCLIP mainly from three perspectives: (1) We propose a Causal Understanding Module to reinforce the model's capability for environmental understanding; (2) We implement a Multi-modal transformer for feature encoding; (3) We fine-tune the CLIP visual encoder in the training process instead of holding it fixed. Therefore, we conduct ablation studies to examine the impact of these components. We conduct the studies on all three tasks and present the result in Table~\ref{tab:ablation_rst}. 


\subsubsection{Impact of Causal Understanding Module}
\label{exp_ablation_causalrnn}
We first examine the impact of the Causal Understanding Module by implementing this module on EmbCLIP \cite{khandelwal2022simple}. We keep all the configurations in EmbCLIP unchanged and compare the cases with/without the causal understanding module. We refer to this model as \textbf{Causal-RNN} in Table~\ref{tab:ablation_rst} for simplicity.

\begin{wrapfigure}{R}{0.6\textwidth}
    \vspace{-20pt}
  \centering
  \includegraphics[width=1\linewidth]{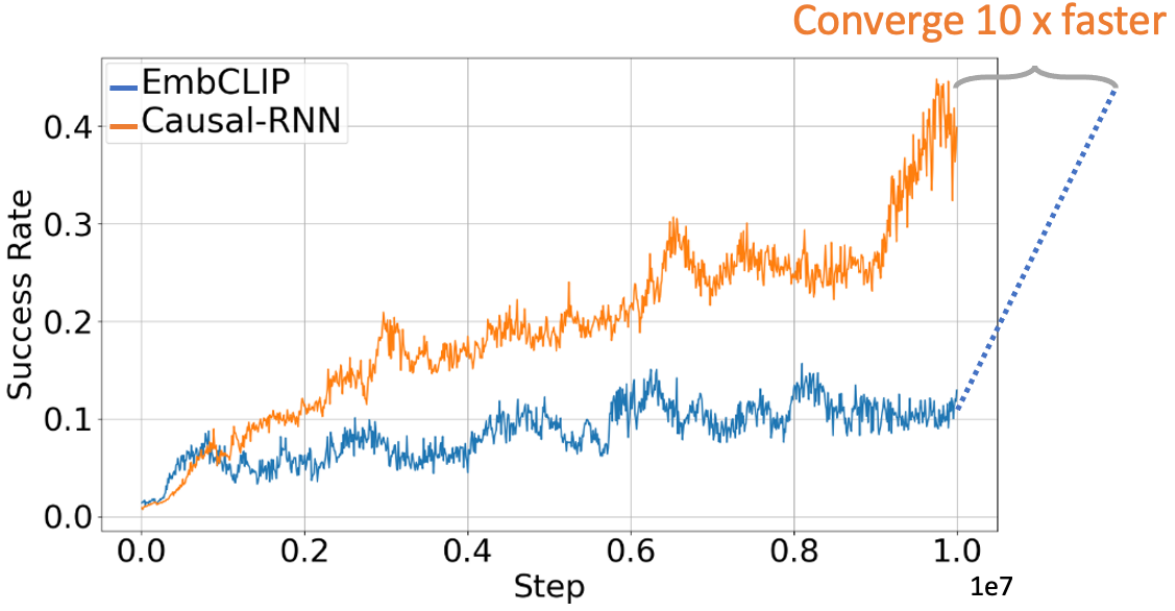}
  \caption{Average Success Rate for EmbCLIP and Causal-RNN on RoboTHOR ObjNav. Our proposed Causal Understanding Module can: 1) significantly benefit the performance; and 2) significantly reduce the training time by 10 times.}
  \label{fig:ablation1_sr_curve}
  \vspace{-13pt}
\end{wrapfigure}

Besides, we also plot a curve of Success Rate in Figure~\ref{fig:ablation1_sr_curve} for direct comparison of the training progress, where the x-axis denotes the steps of training, and the y-axis denotes the Success Rate. We ran each case 10 times randomly and plotted the average Success Rate at each step. Due to the space limitation, we only plotted the curve for the first 10M steps.

We observe two clear benefits of implementing the Causal Understanding Module: (1) It significantly improves the performance of the baseline model without computational overhead. Specifically, we improve the success rate by more than 50\% by adding one simple linear layer; (2) It significantly reduces the time required for training a satisfactory agent. Specifically, we trained the Causal-RNN model for 20M steps and achieved a success rate of 0.48. In contrast, as claimed in \cite{khandelwal2022simple}, EmbCLIP achieved a success rate of 0.47 after training for 200M steps. Therefore, we significantly reduce the training time 10 times by adding one simple Linear layer.


\begin{table*}[t]
  \centering
  \caption{Result of Ablation Studies. Fine-tuning CLIP and implementing the Transformer Encoder benefit the performance, but the performance gain of our method mainly comes from the Causal Understanding Module.}
  \setlength{\tabcolsep}{4pt} 
  \renewcommand{\arraystretch}{1.2} 
  \small
  \begin{tabular}{l | l l | l l l | l l l}
    \toprule
    & \multicolumn{2}{c}{Robo ObjNav} & \multicolumn{3}{c}{Habitat ObjNav} & \multicolumn{3}{c}{Habitat PointNav}\\
    & SPL $\uparrow$ & SR $\uparrow$ & SPL $\uparrow$ & SR $\uparrow$ & GD $\downarrow$ & SPL $\uparrow$ & SR $\uparrow$ & GD $\downarrow$ \\
    
    \midrule
    
    EmbCLIP & 0.14 & 0.31 & 0.07 & 0.15 & 7.13 & 0.80 & 0.92 & 0.48 \\
    
    \midrule
    
    EmbCLIP (f.t.) & 0.20  & 0.41 & 0.09 & 0.21 & 7.07  & 0.84  & 0.94  & 0.43  \\
    
    Transformer & 0.18  & 0.36 & 0.08 & 0.18  & 7.08 & 0.83  & 0.93  & 0.45 \\
    
    Causal-RNN & 0.23  & 0.48 & 0.11 & 0.25 & 7.01  & 0.88  & 0.96 & 0.40 \\
    
    \midrule

    CAT (Ours) & 0.31 & 0.69 & 0.16 & 0.41  & 6.76  & 0.93  & 0.98  & 0.34 \\
    
    \bottomrule
  \end{tabular}
  \label{tab:ablation_rst}
\end{table*}

\subsubsection{Impact of Multi-modal Transformer}
To examine the effectiveness of the Multi-modal Transformer Encoder, we simplify the architecture by removing the Causal Understanding Module from Figure~\ref{fig:framework}, and holding CLIP fixed across the training process while keeping all other settings unchanged. In other words, we replace the encoder in EmbCLIP with our Multi-modal Transformer. We directly compare this method with EmbCLIP to understand the benefits of employing a multi-modal Transformer. We refer to this method as \textbf{Transformer} in Table~\ref{tab:ablation_rst}, and observe that the Multi-modal Transformer Encoder is beneficial to the model's performance.

\subsubsection{Impact of CLIP Fine-tuning}
\label{exp_ablation_clipfinetune} 
To examine the impact of a tunable visual encoder, we modify the architecture of EmbCLIP slightly by making the CLIP visual encoder trainable, and all other settings remain unchanged. We refer to this method as \textbf{EmbCLIP (f.t.)} in Table~\ref{tab:ablation_rst}, and find that a tunable visual encoder is beneficial to the outcome.

\subsubsection{Summary} Our ablation studies reveal that all three components are necessary for our method to achieve optimal outcomes. However, while a Multi-modal Transformer and a tunable Visual Encoder contribute to the agent's performance, the majority of the observed improvements in Section~\ref{exp_rst} can be attributed to the Causal Understanding Module.

\subsection{Causal Understanding Module in Supervised Learning}
\label{exp_other_setting}

While our primary focus in the earlier sections is on experiments within the Reinforcement Learning setting, we also assess the effectiveness of our proposed method in Supervised Learning. Specifically, we implemented the Causal Understanding Module on existing methods including Seq2Seq \cite{anderson2018vision}, Speaker Follower \cite{fried2018speaker} and EnvDrop \cite{tan2019learning}, while maintaining all other settings as proposed in the respective papers. The experiments are conducted on the R2R dataset \cite{anderson2018vision}, and we evaluate the performance of the methods by the commonly used metrics for the task including Navigation Error (NE), Oracle Success Rate (OSR), and SR introduced in earlier sections. The results of the validation unseen split are presented in Table~\ref{tab:rst_r2r}. Overall, the results have demonstrated the effectiveness, consistency and generalizability of the proposed Causal Understanding Module.

\begin{table}
  \centering
  \setlength{\tabcolsep}{5pt}
  \small
  \vspace{-5pt}
  \caption{Causal Understanding Module in Supervised Learning.}
  \begin{tabular}{l | l l l}
    \toprule
    & NE $\downarrow$ & OSR $\uparrow$ & SR $\uparrow$\\
    \midrule
    Seq2Seq & 7.8 & 28.4 & 21.8 \\
    \:\:\: + Ours & 5.8 \textcolor{teal}{$\downarrow$(-2.0)} & 38.1 \textcolor{teal}{$\uparrow$(+9.7)} & 41.3 \textcolor{teal}{$\uparrow$(+19.5)} \\
    \midrule
    Speaker Follower & 6.6 & 45.0 & 35.0 \\
    \:\:\: + Ours  & 4.9 \textcolor{teal}{$\downarrow$(-1.7)} & 53.1 \textcolor{teal}{$\uparrow$(+8.1)} & 52.4 \textcolor{teal}{$\uparrow$(+17.4)} \\
    \midrule
    EnvDrop & 5.2 & - & 52.2 \\
    \:\:\: + Ours & 4.5 \textcolor{teal}{$\downarrow$(-0.7)} & - & 68.7 \textcolor{teal}{$\uparrow$(+16.5)} \\
    \bottomrule
  \end{tabular}
  \label{tab:rst_r2r}
  \vspace{-10pt}
\end{table}

\subsection{Computational Cost}
Our proposed architecture requires more computational resources than the baseline method. However, our experiments reveal that our method is not computationally expensive. We trained the model on our machine with a single NVIDIA TITAN X GPU, and it occupies only 5GB of Memory to train the model. It takes around 40 GPU hours to train the model for 20M steps, which is nearly identical to the baseline (EmbCLIP). Besides, our Causal-RNN model in Section~\ref{exp_ablation_causalrnn} is proven to be effective without any extra resources.

\section{Related Work}
Many simulators and tasks have been proposed in recent years for Embodied AI \cite{deitke2020robothor,shen2021igibson,szot2021habitat,xia2018gibson,batra2020objectnav,weihs2021visual,shridhar2020alfred}. For navigation tasks\cite{gadre2023cows}, while fast progress has been made in recent years, most methods are specifically designed for certain tasks aiming for better results for the challenges \cite{morin2023one,kareer2023vinl,yu2023frontier,zheng2022jarvis,min2021film,inoue2022prompter,blukis2022persistent,zhang2021hierarchical,suglia2021embodied,deitke2022️,pashevich2021episodic}. However, although these methods achieve state-of-the-art performances, they cannot be generalized to other scenarios due to the dataset-specific architecture, inductive bias and hand-crafted logic \cite{duan2022survey,khandelwal2022simple}. Recently, \cite{khandelwal2022simple} addressed these issues. They found that CLIP \cite{radford2021learning} makes an effective visual encoder and proposed a novel method that does not require any task-specific architectures and inductive bias. Besides, some works also use a Transformer Encoder in the agent due to its strong capability \cite{moudgil2021soat,wang2024navformer}. Recent studies have also focused on training foundation models to enhance the performance of agents in Navigation tasks \cite{shah2023vint}.

\section{Conclusion}
In conclusion, this paper addresses key challenges in the domain of Navigation tasks, particularly the limitations associated with prevalent vanilla sequential data modelling methods and task-specific designs, which often hinder generalizability and performance. By elucidating the intrinsic disparities between Navigation tasks and conventional sequential data modelling tasks, we introduce a novel causal framework to explain the necessity of the causal environment understanding module and proposed Causality-Aware Transformer (CAT), an End-to-End transformer-based method that exhibits notable performance improvements across diverse tasks and simulators, surpassing baseline approaches on multiple benchmarks. Furthermore, comprehensive ablation studies reveal that most of the performance gain of our method can be attributed to the Causal Understanding Module, which is proven to be effective and can be implemented in other methods across the paradigm of Reinforcement Learning and Supervised Learning without computational overhead.

\bibliographystyle{splncs04}
\bibliography{9780}

\begin{thebibliography}{10}
\providecommand{\url}[1]{\texttt{#1}}
\providecommand{\urlprefix}{URL }
\providecommand{\doi}[1]{https://doi.org/#1}

\bibitem{anderson2018vision}
Anderson, P., Wu, Q., Teney, D., Bruce, J., Johnson, M., S{\"u}nderhauf, N., Reid, I., Gould, S., Van Den~Hengel, A.: Vision-and-language navigation: Interpreting visually-grounded navigation instructions in real environments. In: Proceedings of the IEEE conference on computer vision and pattern recognition. pp. 3674--3683 (2018)

\bibitem{batra2020objectnav}
Batra, D., Gokaslan, A., Kembhavi, A., Maksymets, O., Mottaghi, R., Savva, M., Toshev, A., Wijmans, E.: Objectnav revisited: On evaluation of embodied agents navigating to objects. arXiv preprint arXiv:2006.13171  (2020)

\bibitem{blukis2022persistent}
Blukis, V., Paxton, C., Fox, D., Garg, A., Artzi, Y.: A persistent spatial semantic representation for high-level natural language instruction execution. In: Conference on Robot Learning. pp. 706--717. PMLR (2022)

\bibitem{chang2017matterport3d}
Chang, A., Dai, A., Funkhouser, T., Halber, M., Niessner, M., Savva, M., Song, S., Zeng, A., Zhang, Y.: Matterport3d: Learning from rgb-d data in indoor environments. arXiv preprint arXiv:1709.06158  (2017)

\bibitem{deitke2020robothor}
Deitke, M., Han, W., Herrasti, A., Kembhavi, A., Kolve, E., Mottaghi, R., Salvador, J., Schwenk, D., VanderBilt, E., Wallingford, M., et~al.: Robothor: An open simulation-to-real embodied ai platform. In: Proceedings of the IEEE/CVF conference on computer vision and pattern recognition. pp. 3164--3174 (2020)

\bibitem{deitke2022️}
Deitke, M., VanderBilt, E., Herrasti, A., Weihs, L., Ehsani, K., Salvador, J., Han, W., Kolve, E., Kembhavi, A., Mottaghi, R.: Procthor: Large-scale embodied ai using procedural generation. NeurIPS  \textbf{35},  5982--5994 (2022)

\bibitem{duan2022survey}
Duan, J., Yu, S., Tan, H.L., Zhu, H., Tan, C.: A survey of embodied ai: From simulators to research tasks. IEEE Transactions on Emerging Topics in Computational Intelligence  \textbf{6}(2),  230--244 (2022)

\bibitem{francis2022core}
Francis, J., Kitamura, N., Labelle, F., Lu, X., Navarro, I., Oh, J.: Core challenges in embodied vision-language planning. Journal of Artificial Intelligence Research  \textbf{74},  459--515 (2022)

\bibitem{fried2018speaker}
Fried, D., Hu, R., Cirik, V., Rohrbach, A., Andreas, J., Morency, L.P., Berg-Kirkpatrick, T., Saenko, K., Klein, D., Darrell, T.: Speaker-follower models for vision-and-language navigation. Advances in neural information processing systems  \textbf{31} (2018)

\bibitem{gadre2023cows}
Gadre, S.Y., Wortsman, M., Ilharco, G., Schmidt, L., Song, S.: Cows on pasture: Baselines and benchmarks for language-driven zero-shot object navigation. In: Proceedings of the IEEE/CVF Conference on Computer Vision and Pattern Recognition. pp. 23171--23181 (2023)

\bibitem{inoue2022prompter}
Inoue, Y., Ohashi, H.: Prompter: Utilizing large language model prompting for a data efficient embodied instruction following. arXiv preprint arXiv:2211.03267  (2022)

\bibitem{kareer2023vinl}
Kareer, S., Yokoyama, N., Batra, D., Ha, S., Truong, J.: Vinl: Visual navigation and locomotion over obstacles. In: 2023 IEEE International Conference on Robotics and Automation (ICRA). pp. 2018--2024. IEEE (2023)

\bibitem{khandelwal2022simple}
Khandelwal, A., Weihs, L., Mottaghi, R., Kembhavi, A.: Simple but effective: Clip embeddings for embodied ai. In: Proceedings of the IEEE/CVF Conference on Computer Vision and Pattern Recognition. pp. 14829--14838 (2022)

\bibitem{kingma2014adam}
Kingma, D.P., Ba, J.: Adam: A method for stochastic optimization. arXiv preprint arXiv:1412.6980  (2014)

\bibitem{luo2022stubborn}
Luo, H., Yue, A., Hong, Z.W., Agrawal, P.: Stubborn: A strong baseline for indoor object navigation. In: 2022 IEEE/RSJ International Conference on Intelligent Robots and Systems (IROS). pp. 3287--3293. IEEE (2022)

\bibitem{maksymets2021thda}
Maksymets, O., Cartillier, V., Gokaslan, A., Wijmans, E., Galuba, W., Lee, S., Batra, D.: Thda: Treasure hunt data augmentation for semantic navigation. In: Proceedings of the IEEE/CVF International Conference on Computer Vision. pp. 15374--15383 (2021)

\bibitem{min2021film}
Min, S.Y., Chaplot, D.S., Ravikumar, P., Bisk, Y., Salakhutdinov, R.: Film: Following instructions in language with modular methods. arXiv preprint arXiv:2110.07342  (2021)

\bibitem{morin2023one}
Morin, S., Saavedra-Ruiz, M., Paull, L.: One-4-all: Neural potential fields for embodied navigation. In: 2023 IEEE/RSJ International Conference on Intelligent Robots and Systems (IROS). IEEE (2023)

\bibitem{moudgil2021soat}
Moudgil, A., Majumdar, A., Agrawal, H., Lee, S., Batra, D.: Soat: A scene-and object-aware transformer for vision-and-language navigation. Advances in Neural Information Processing Systems  \textbf{34},  7357--7367 (2021)

\bibitem{pashevich2021episodic}
Pashevich, A., Schmid, C., Sun, C.: Episodic transformer for vision-and-language navigation. In: Proceedings of the IEEE/CVF International Conference on Computer Vision. pp. 15942--15952 (2021)

\bibitem{puterman1990markov}
Puterman, M.L.: Markov decision processes. Handbooks in operations research and management science  \textbf{2},  331--434 (1990)

\bibitem{radford2021learning}
Radford, A., Kim, J.W., Hallacy, C., Ramesh, A., Goh, G., Agarwal, S., Sastry, G., Askell, A., Mishkin, P., Clark, J., et~al.: Learning transferable visual models from natural language supervision. In: International conference on machine learning. pp. 8748--8763. PMLR (2021)

\bibitem{ramrakhya2023pirlnav}
Ramrakhya, R., Batra, D., Wijmans, E., Das, A.: Pirlnav: Pretraining with imitation and rl finetuning for objectnav. In: Proceedings of the IEEE/CVF Conference on Computer Vision and Pattern Recognition. pp. 17896--17906 (2023)

\bibitem{ramrakhya2022habitat}
Ramrakhya, R., Undersander, E., Batra, D., Das, A.: Habitat-web: Learning embodied object-search strategies from human demonstrations at scale. In: Proceedings of the IEEE/CVF Conference on Computer Vision and Pattern Recognition. pp. 5173--5183 (2022)

\bibitem{schulman2017proximal}
Schulman, J., Wolski, F., Dhariwal, P., Radford, A., Klimov, O.: Proximal policy optimization algorithms. arXiv preprint arXiv:1707.06347  (2017)

\bibitem{shah2023vint}
Shah, D., Sridhar, A., Dashora, N., Stachowicz, K., Black, K., Hirose, N., Levine, S.: Vint: A foundation model for visual navigation. arXiv preprint arXiv:2306.14846  (2023)

\bibitem{shen2021igibson}
Shen, B., Xia, F., Li, C., Mart{\'\i}n-Mart{\'\i}n, R., Fan, L., Wang, G., P{\'e}rez-D’Arpino, C., Buch, S., Srivastava, S., Tchapmi, L., et~al.: igibson 1.0: A simulation environment for interactive tasks in large realistic scenes. In: 2021 IEEE/RSJ International Conference on Intelligent Robots and Systems (IROS). pp. 7520--7527. IEEE (2021)

\bibitem{shridhar2020alfred}
Shridhar, M., Thomason, J., Gordon, D., Bisk, Y., Han, W., Mottaghi, R., Zettlemoyer, L., Fox, D.: Alfred: A benchmark for interpreting grounded instructions for everyday tasks. In: Proceedings of the IEEE/CVF conference on computer vision and pattern recognition. pp. 10740--10749 (2020)

\bibitem{smith2005development}
Smith, L., Gasser, M.: The development of embodied cognition: Six lessons from babies. Artificial life  \textbf{11}(1-2),  13--29 (2005)

\bibitem{suglia2021embodied}
Suglia, A., Gao, Q., Thomason, J., Thattai, G., Sukhatme, G.: Embodied bert: A transformer model for embodied, language-guided visual task completion. arXiv preprint arXiv:2108.04927  (2021)

\bibitem{szot2021habitat}
Szot, A., Clegg, A., Undersander, E., Wijmans, E., Zhao, Y., Turner, J., Maestre, N., Mukadam, M., Chaplot, D.S., Maksymets, O., et~al.: Habitat 2.0: Training home assistants to rearrange their habitat. Advances in Neural Information Processing Systems  \textbf{34},  251--266 (2021)

\bibitem{tan2019learning}
Tan, H., Yu, L., Bansal, M.: Learning to navigate unseen environments: Back translation with environmental dropout. arXiv preprint arXiv:1904.04195  (2019)

\bibitem{vaswani2017attention}
Vaswani, A., Shazeer, N., Parmar, N., Uszkoreit, J., Jones, L., Gomez, A.N., Kaiser, {\L}., Polosukhin, I.: Attention is all you need. Advances in neural information processing systems  \textbf{30} (2017)

\bibitem{wang2024navformer}
Wang, H., Tan, A.H., Nejat, G.: Navformer: A transformer architecture for robot target-driven navigation in unknown and dynamic environments. IEEE Robotics and Automation Letters  (2024)

\bibitem{weihs2021visual}
Weihs, L., Deitke, M., Kembhavi, A., Mottaghi, R.: Visual room rearrangement. In: Proceedings of the IEEE/CVF conference on computer vision and pattern recognition. pp. 5922--5931 (2021)

\bibitem{weihs2020allenact}
Weihs, L., Salvador, J., Kotar, K., Jain, U., Zeng, K.H., Mottaghi, R., Kembhavi, A.: Allenact: A framework for embodied ai research. arXiv preprint arXiv:2008.12760  (2020)

\bibitem{wijmans2019dd}
Wijmans, E., Kadian, A., Morcos, A., Lee, S., Essa, I., Parikh, D., Savva, M., Batra, D.: Dd-ppo: Learning near-perfect pointgoal navigators from 2.5 billion frames. arXiv preprint arXiv:1911.00357  (2019)

\bibitem{xia2018gibson}
Xia, F., Zamir, A.R., He, Z., Sax, A., Malik, J., Savarese, S.: Gibson env: Real-world perception for embodied agents. In: Proceedings of the IEEE conference on computer vision and pattern recognition. pp. 9068--9079 (2018)

\bibitem{ye2021auxiliary}
Ye, J., Batra, D., Das, A., Wijmans, E.: Auxiliary tasks and exploration enable objectgoal navigation. In: Proceedings of the IEEE/CVF International Conference on Computer Vision. pp. 16117--16126 (2021)

\bibitem{yu2023frontier}
Yu, B., Kasaei, H., Cao, M.: Frontier semantic exploration for visual target navigation. In: 2023 IEEE International Conference on Robotics and Automation (ICRA). IEEE (2023)

\bibitem{zhang2021hierarchical}
Zhang, Y., Chai, J.: Hierarchical task learning from language instructions with unified transformers and self-monitoring. arXiv preprint arXiv:2106.03427  (2021)

\bibitem{zheng2022jarvis}
Zheng, K., Zhou, K., Gu, J., Fan, Y., Wang, J., Di, Z., He, X., Wang, X.E.: Jarvis: A neuro-symbolic commonsense reasoning framework for conversational embodied agents. arXiv preprint arXiv:2208.13266  (2022)

\end{thebibliography}

\end{document}